\documentclass[letterpaper, 10 pt, conference]{ieeeconf}  % Comment this line out if you need a4paper

\IEEEoverridecommandlockouts                              % This command is only needed if 
\usepackage{graphics} % for pdf, bitmapped graphics files
\usepackage{epsfig} % for postscript graphics files
\usepackage{mathptmx} % assumes new font selection scheme installed
\usepackage{times} % assumes new font selection scheme installed
\usepackage{amsmath} % assumes amsmath package installed
\usepackage{amssymb}  % assumes amsmath package installed
\usepackage{gensymb}
\usepackage{graphicx}
\usepackage{algorithm}
\usepackage{algorithmic}
\usepackage{caption}
\usepackage{hyperref}

\hypersetup{
    colorlinks=true,
    linkcolor=blue,
    filecolor=magenta,      
    urlcolor=blue,
    pdftitle={CAROM Air},
    pdfpagemode=FullScreen,
}

\title{\LARGE \bf
CAROM Air - Vehicle Localization and Traffic Scene Reconstruction from Aerial Videos
}

\author{Duo Lu$^{1}$, Eric Eaton$^{1}$, Matt Weg$^{1}$, Wei Wang$^{2}$, Steven Como$^{2}$, Jeffrey Wishart$^{2}$, Hongbin Yu$^{2}$, Yezhou Yang$^{2}$%
\thanks{$^{1}$D. Lu, E. Eaton, and M. Weg are with Rider University.
{\tt\small \{dlu, eatone, wegma\}@rider.edu}}%
\thanks{$^{2}$W. Wang, S. Como, J. Wishart, H. Yu, and Y. Yang are with Arizona State University.
{\tt\small \{wwang266, scomo, jeffrey.wishart, Hongbin.Yu, yz.yang\}@asu.edu}}%
\thanks{This research is sponsored by Rider University Faculty Research Fellowship, MacMillan Scientific Research Fellowship, and the Institute of Automated Mobility (Arizona). We thank Arizona DOT, Greg Leeming, Marisa Paula Walker, Larry Head, Varun Chandra Jammula, Lu Zhao, Baihan Chen, Ke Fan, Nannan Zhang, Zhichao Li for their help.}% <-this % stops a space
}

\begin{document}

\maketitle
\thispagestyle{empty}
\pagestyle{empty}

%%%%%%%%%%%%%%%%%%%%%%%%%%%%%%%%%%%%%%%%%%%%%%%%%%%%%%%%%%%%%%%%%%%%%%%%%%%%%%%%
\begin{abstract}

Road traffic scene reconstruction from videos has been desirable by road safety regulators, city planners, researchers, and autonomous driving technology developers. However, it is expensive and unnecessary to cover every mile of the road with cameras mounted on the road infrastructure. This paper presents a method that can process aerial videos to vehicle trajectory data so that a traffic scene can be automatically reconstructed and accurately re-simulated using computers. On average, the vehicle localization error is about 0.1 m to 0.3 m using a consumer-grade drone flying at 120 meters. This project also compiles a dataset of 50 reconstructed road traffic scenes from about 100 hours of aerial videos to enable various downstream traffic analysis applications and facilitate further road traffic related research. The dataset is available at \url{https://github.com/duolu/CAROM}.

\end{abstract}

%%%%%%%%%%%%%%%%%%%%%%%%%%%%%%%%%%%%%%%%%%%%%%%%%%%%%%%%%%%%%%%%%%%%%%%%%%%%%%%%
\section{Introduction}

Road traffic has created many problems that need to be studied with real-world road traffic data. For example, local Departments of Transportation (DOTs) need to count the vehicles on every major road segment for traffic management purposes. It is desirable that each counted vehicle in the data can have fine-grained attributes, such as vehicle type, speed, lane, etc. City planners and transportation system engineers also want to use detailed road traffic data for better decision-making and resource provisioning. Additionally, for researchers and regulators interested in road safety analysis and driver behavior modeling, it is more valuable to capture the comprehensive motion states of vehicles passing through a specific traffic scene instead of obtaining just a count (which does not carry much information) or a crash report (which happens infrequently). For example, aggressive lane switches and frequent close call incidents on a highway segment may indicate that the traffic is reaching the designed capacity. Similarly, reckless driving behaviors can reveal more insights on road safety than reported accidents. Besides the policy makers, vehicle manufacturers and insurance companies can also benefit from datasets of vehicle trajectories, especially if such data can be used to accurately reconstruct and re-simulate the captured traffic scenes.

\begin{figure}[]
    % \centering
    \begin{center}
        \includegraphics[width=3.4in]{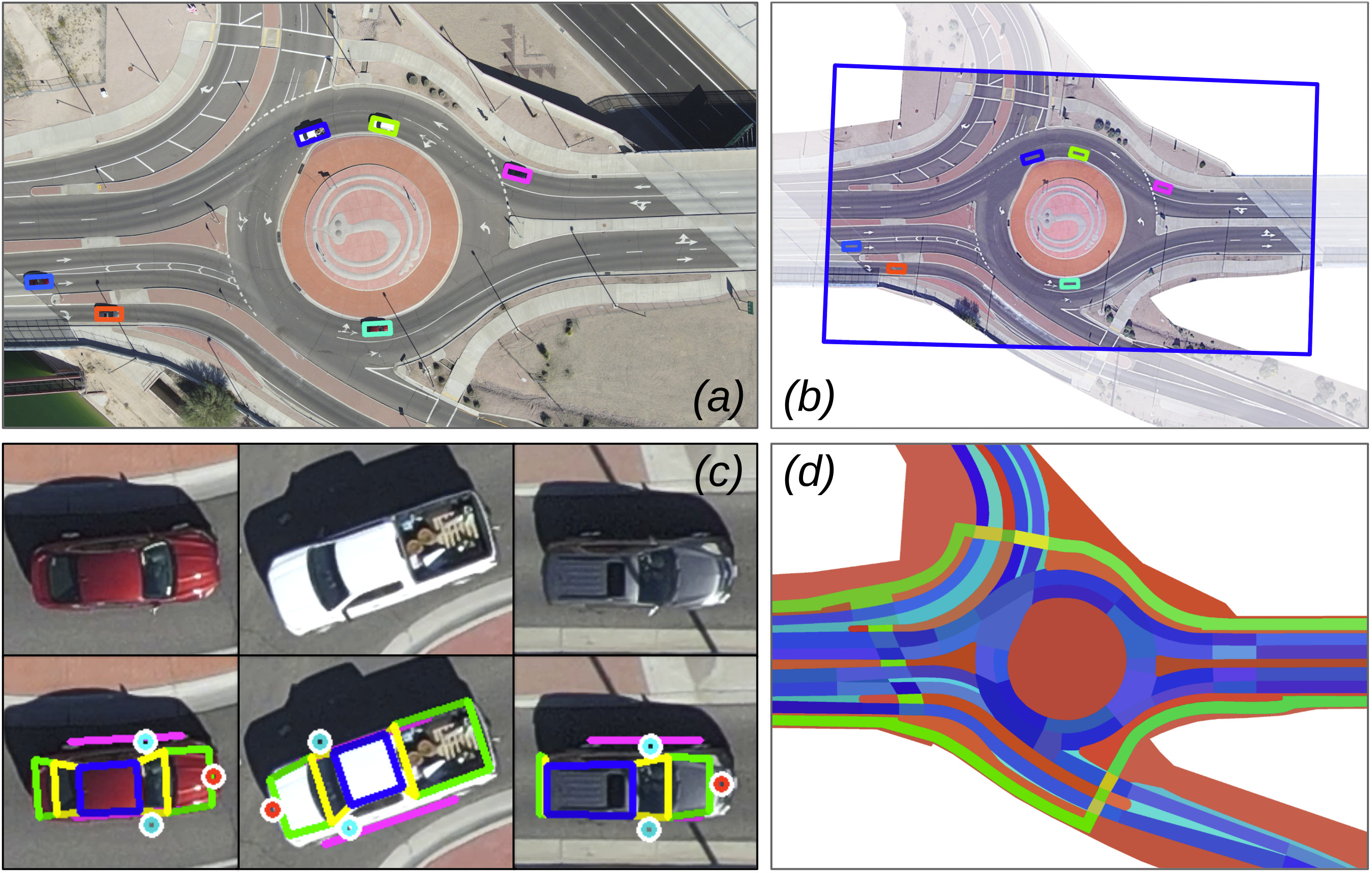}
    \caption{An overview of CAROM Air: (a) tracked vehicles on the aerial video, (b) the reconstructed traffic scene, (c) vehicle keypoints, and (d) the map with semantic annotation.}
    \label{fig:overview}
    \end{center}
\vspace{-0.3in}
\end{figure}

Traditionally, such road traffic data is collected and managed by the DOTs using devices installed on the road infrastructure. There are a few problems. First, it is expensive and unnecessary to cover every mile of the road with sensors and cameras. As a result, many interesting traffic scenes are not captured. Second, although there are many cameras deployed in strategic locations in major cities, it is challenging to process the videos or deliver them over the network due to the sheer volume of the videos. Hence, the operational cost of these cameras further hinders large-scale deployment. Meanwhile, since the cameras cannot move, the captured video data contain redundant information from repeated patterns. Third, the vehicle localization accuracy of infrastructure-based sensors can significantly degrade when a tracked vehicle is far away or occluded by other vehicles. Fourth, due to regulations and privacy issues, it is challenging for researchers outside the DOTs and industrial partners to access these video data for open research purposes. On the other hand, for independent researchers or companies, it is expensive to collect and manage road traffic data. Last, for the researchers and companies who can afford to collect data on the road using cameras and LIDAR sensors, it is time-consuming to label the data to train neural network models for vehicle detection and tracking. This is particularly an issue if the labeling must be done in the 3D space.

To address these issues, we propose a framework named CAROM Air (``\underline{CAR}s \underline{O}n the \underline{M}ap tracked from the \underline{Air}''), which digitizes and reconstructs road traffic scenes from aerial videos taken by drones, as shown in Fig. \ref{fig:overview}. It is inexpensive and flexible since it does not require any support from road infrastructure. The core of this framework is a pipeline that can track vehicles on the aerial videos and localize them on the map accurately through the detection of vehicle keypoints. This allows us to convert the aerial videos to vehicle trajectory data which can be delivered over communication networks for reconstruction or further analysis using programs. Such vehicle trajectory data does not have any personal identifiable information, and hence, they can be shared without causing privacy issues. Moreover, we demonstrate that our data can be used as reference measurements or 3D labels for videos and LIDAR point clouds captured by devices on the road infrastructure. This work is a continuation of the ongoing research conducted by the Institute of Automated Mobility (IAM) \cite{IAM} to support the development and validation of an operational safety assessment methodology \cite{wishart2020driving}\cite{jammula2022evaluation}\cite{kidambi2022sensitivity}\cite{das2023comparison} and intelligent road traffic infrastructure \cite{lu2021carom}\cite{altekar2021infrastructure}\cite{srinivasan2022infrastructure}. In summary, our contributions are as follows.

%It runs an online algorithm that could be deployed on the drone given enough computing resources. 

\begin{enumerate}

\item We developed a keypoint-based vehicle tracking and localization pipeline for aerial videos. The average vehicle localization error is from 0.1 m to 0.3 m using a drone flying at 120 meters in various conditions.
\item We built a dataset of vehicle trajectories obtained from about 100 hours of drone video in 50 different road traffic scenes. 
\item For each scene, we also provide the map with semantic segmentation at the lane level, which enables further automated traffic analysis and statistics.
\item We demonstrated several downstream applications to show the practicality of our framework.

\end{enumerate}

% The remainder of this paper is organized as follows. Section 2 discusses related work, and section 3 presents our proposed framework. After that, section 4 details our dataset, section 5 shows our experimental evaluation, and section 6 shows example applications. Finally, section 7 draws conclusions and discusses future work.

\section{Related Work}

Unmanned Aerial Vehicles (UAVs), commonly called drones, have been used in 3D mapping of road infrastructures \cite{nex2014uav}, traffic monitoring \cite{butilua2022urban}\cite{guirado2021stratotrans}\cite{gupta2022monitoring}\cite{christodoulou2020optimized}\cite{barmpounakis2020new}\cite{niu2018uav}\cite{lee2015examining}\cite{khan2020smart}\cite{krajewski2021drone}, road safety analysis \cite{outay2020applications}, and transportation of humans or goods \cite{gupta2021advances}. They are gaining popularity as an inexpensive and flexible method of obtaining aerial videos of road traffic scenes. To further process the videos, a pipeline of vehicle detection and tracking can be applied, typically with deep neural networks \cite{srivastava2021survey}. With such methods, researchers have constructed datasets of vehicle trajectories obtained from drone videos \cite{8569552}\cite{9304839}\cite{9294728}\cite{9827305}\cite{9294301}\cite{zhan2019interaction}\cite{zheng2022citysim}, which supplement existing large-scale autonomous driving datasets \cite{sun2020scalability}\cite{chang2019argoverse}\cite{wilson2021argoverse}\cite{huang2018apolloscape}\cite{geiger2013vision} and road infrastructure based traffic monitoring datasets \cite{tang2019cityflow}\cite{krammer2019providentia}\cite{zou2022real}. These vehicle trajectory datasets further enable a series of analysis tasks, often with the help of a map containing lane-level traffic semantics \cite{poggenhans2018lanelet2}\cite{openDrive}. Compared to existing works, our method provides better vehicle localization accuracy with more rigorous and more extensive evaluations. Besides, thanks to our keypoint-based vehicle localization algorithm, our framework has better flexibility in drone camera angles rather than requiring the camera to always look downward. Similar keypoint-based methods have been explored in 3D reconstruction \cite{tulsiani2015viewpoints}\cite{yu2021small}\cite{abdel2022using}\cite{khorramshahi2019dual}\cite{simoni2021improving}\cite{yang2020multi}\cite{hu2022vehicle} and autonomous driving from a ego vehicle's view point \cite{liu2021autoshape}\cite{li2020rtm3d}\cite{murthy2017reconstructing}\cite{kreiss2021openpifpaf}, but not from a drone camera. Additionally, our dataset is larger and more diverse than similar datasets from existing works.

\section{The CAROM Air Framework}

\begin{figure}[ht]
\vspace{-0.05in}
    % \centering
    \begin{center}
        \includegraphics[width=3.1in]{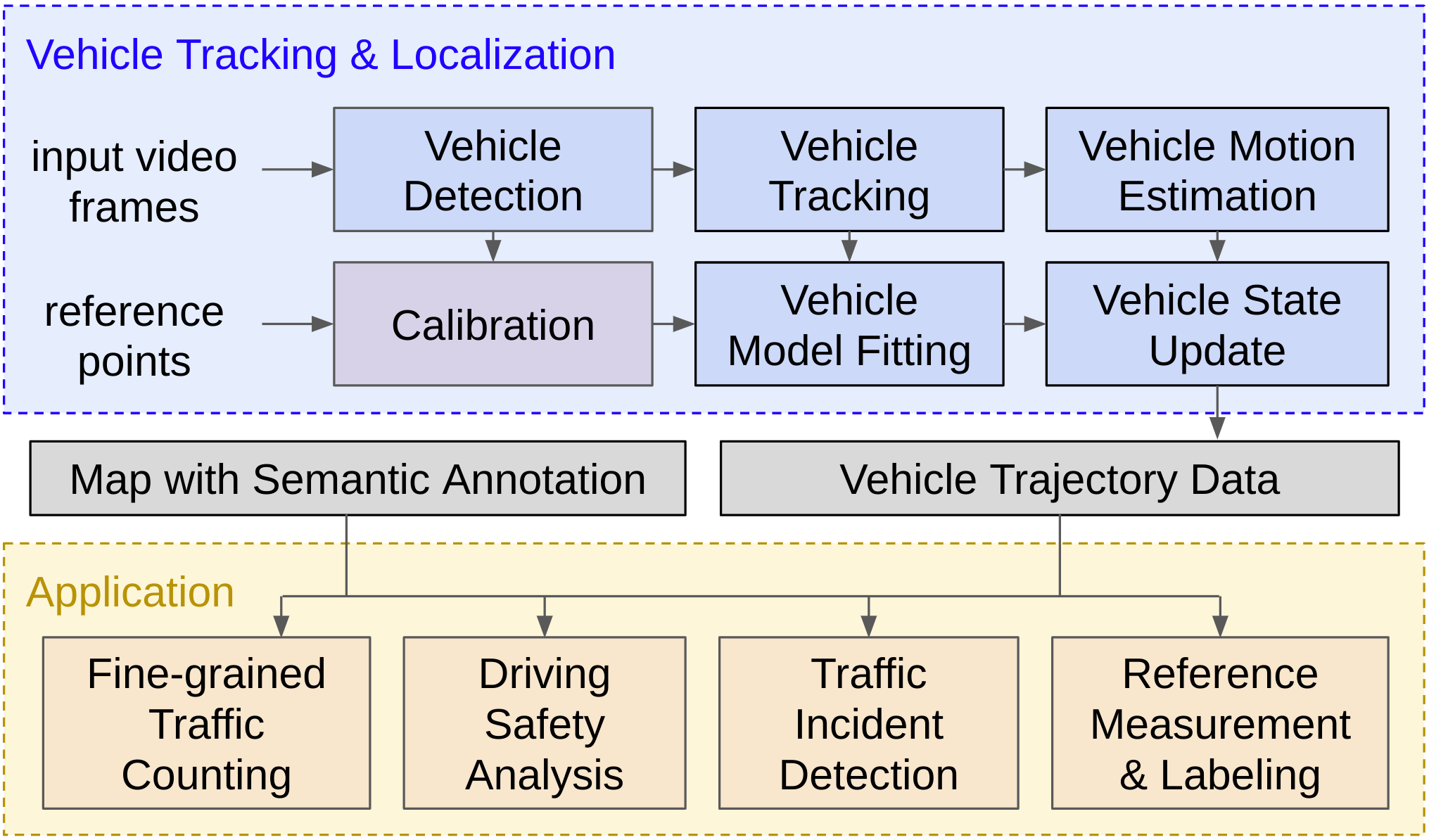}
    \caption{The CAROM Framework Architecture.}
    \label{fig:arch}
    \end{center}
\vspace{-0.2in}
\end{figure}

The CAROM Air framework contains three layers, as shown in Fig. \ref{fig:arch}. The foundational layer is a pipeline to track and localize vehicles captured on the aerial video (detailed in this section). The middle layer is the dataset of tracked vehicle trajectories and traffic scene maps with lane-level semantic annotation (section V). After that, the downstream applications form the third layer (section VI).

\subsection{Camera Calibration}

\begin{figure}[ht]
\vspace{-0.05in}
    % \centering
    \begin{center}
        \includegraphics[width=3.0in]{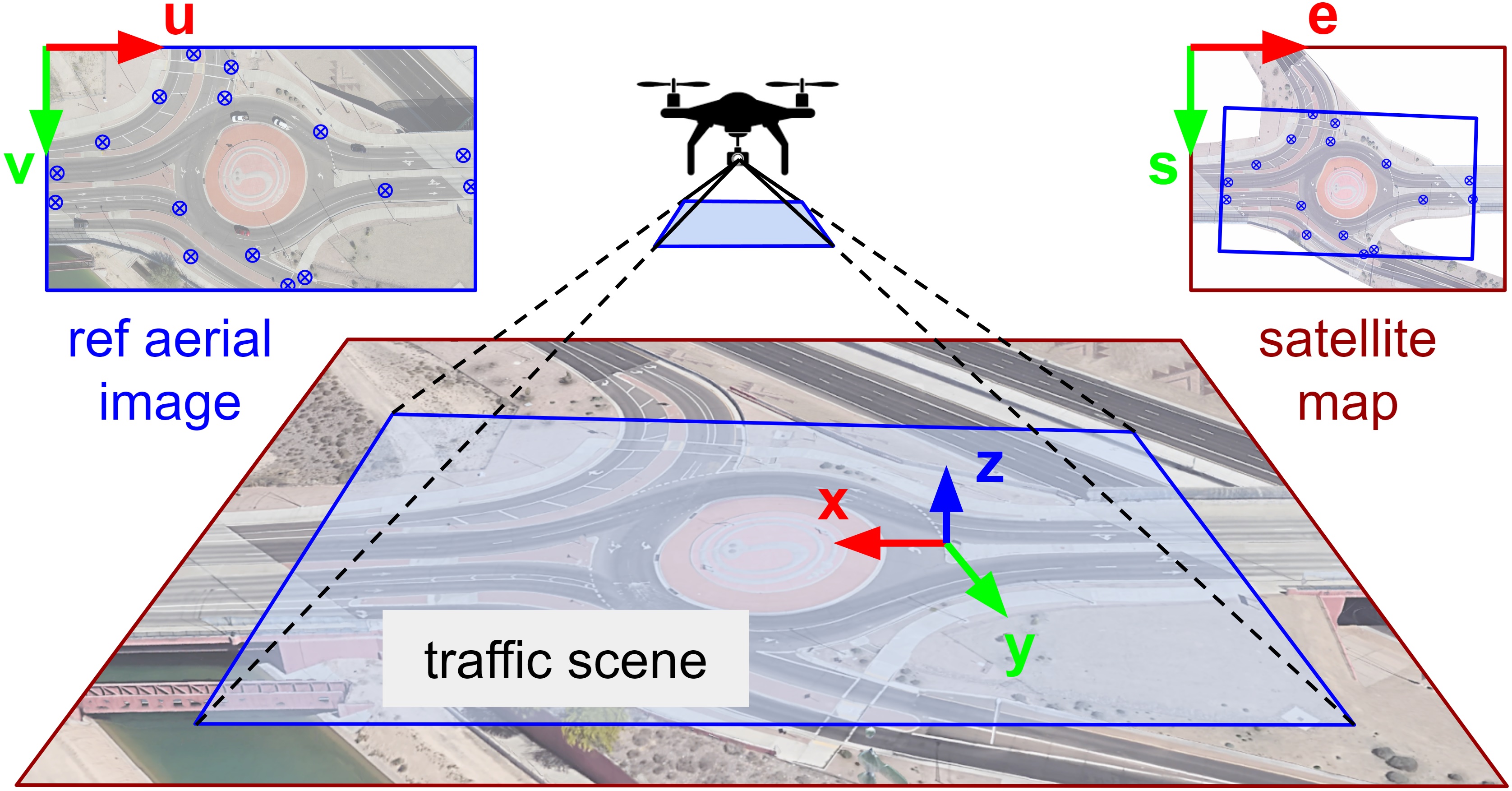}
    \caption{An illustration of camera calibration.}
    \label{fig:calibration}
    \end{center}
\vspace{-0.2in}
\end{figure}

The camera calibration step is illustrated in Fig. \ref{fig:calibration}. We use a pinhole camera model with no distortion and a flat ground model, which can usually achieve enough accuracy. For each video track, we usually annotate 8 to 16 point correspondences on a satellite map (e.g., a screenshot on Google Maps) and a reference aerial image (typically the first image in a video). With these point correspondences, the 3D pose of the camera is solved through Perspective-n-Points (PnP) \cite{hartley2003multiple} given the camera intrinsics (calibrated in the lab) and the 3D coordinates of the points (computed using the scale of the map by assuming the annotated points are on the flat ground). Different from a stationary camera installed on the road infrastructure, the pose of the drone camera can drift. Hence, recalibration is needed for each video image. To achieve this, we detect the corners features \cite{harris1988combined} on the ground (denoted as the reference points in Fig. \ref{fig:arch}) for the reference aerial image, track them across the whole video, and recompute the camera pose using PnP. The semantic annotation of the map helps to determine those points on the ground, e.g., the map shown in Fig. 1(d). With the camera parameters and the map, we can back-project any image pixel to a 3D location if that pixel is on the ground.

\subsection{Vehicle Detection and Tracking}

\begin{figure}[]
\vspace{0.07in}
    % \centering
    \begin{center}
        \includegraphics[width=3.3in]{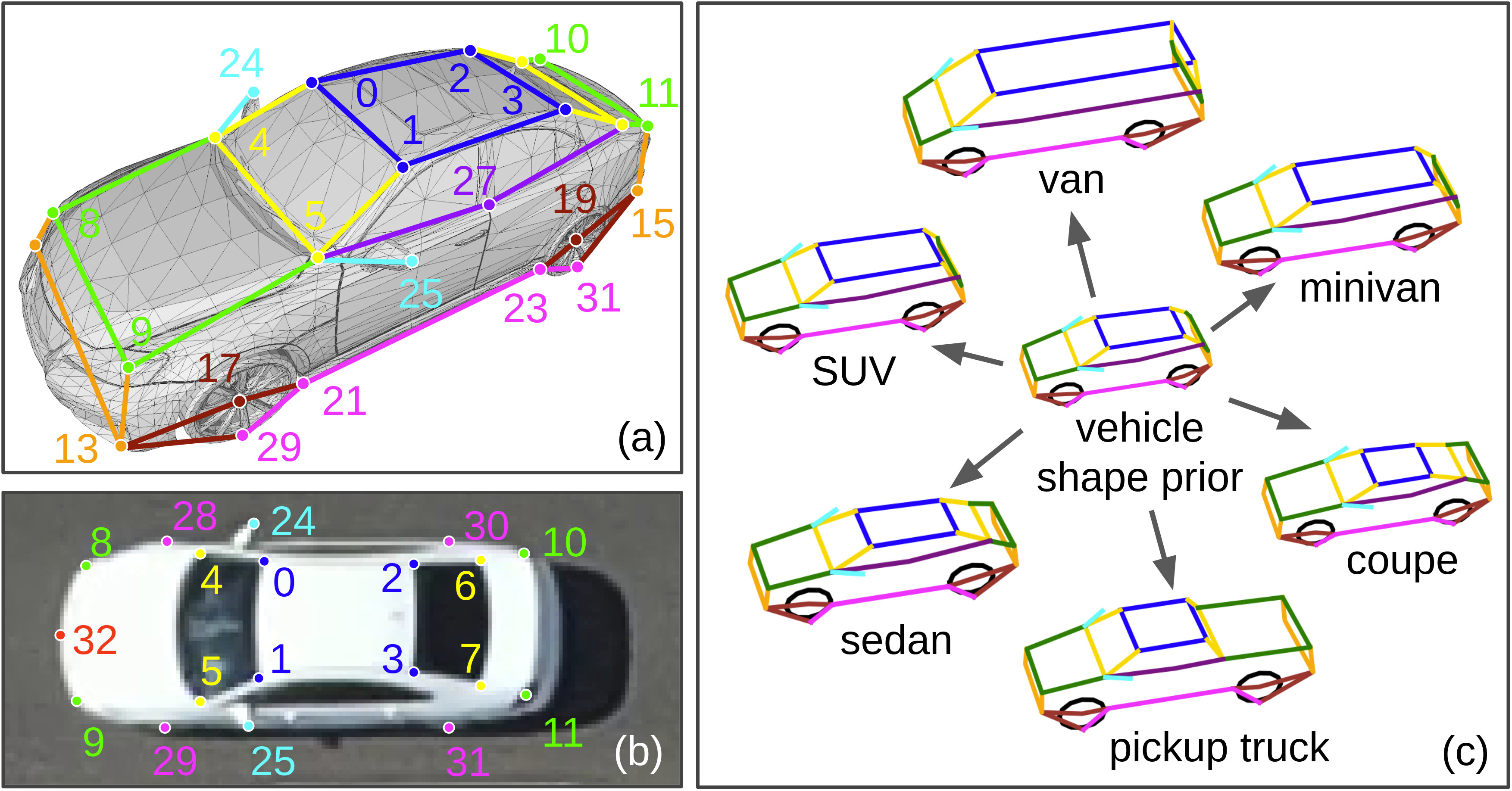}
    \caption{Vehicle keypoints: (a) defined in 3D, (b) detected on the image, and (c) generated from a vehicle shape prior.}
    \label{fig:keypoints}
    \end{center}
\vspace{-0.25in}
\end{figure}

We use a Keypoint RCNN \cite{he2017mask} to detect vehicle keypoints and bounding boxes on each image. We define 33 keypoints in total, as shown in TABLE \ref{tb:kps} and illustrated in Fig. \ref{fig:keypoints} (a). Keypoints are usually defined in groups of two (i.e., right-left) or four (i.e., front-right, front-left, rear-right, and rear-left). Among them, 19 keypoints are detected on the image, as shown in Fig. \ref{fig:keypoints} (b). These keypoints are usually related to observable features such as corners. Hence, they can be reliably detected in most cases using a Keypoint RCNN trained from a small dataset constructed by us (4,386 images, about 12,000 vehicles in total). With the detected vehicle object instances on two adjacent video frames, we associate them if the intersection-over-union (IoU) of their bounding boxes exceeds a certain threshold (i.e., tracking by detection).

\begin{table}[ht]
%\vspace{-0.1in}
\caption{Definition of vehicle keypoints.}
\label{tb:kps}
\centering
\small
\begin{tabular}{|c|c|l|}
\hline
ID  &  detected?  & keypoint definition             \\ \hline \hline
0 - 3   & Yes & corners of roof top                         \\ \hline
4 - 7   & Yes & corners of font and rear windshields         \\ \hline
8 - 11  & Yes & centers of front and rear lights            \\ \hline
12 - 15 & No  & corners of front and rear bumpers           \\ \hline
16 - 19 & No  & centers of wheels                           \\ \hline
20 - 23 & No  & corners of chassis bottom surface           \\ \hline
24 - 25 & Yes & outermost corners of side mirrors           \\ \hline
26 - 27 & No  & corners of the front door windows           \\ \hline
28 - 31 & Yes & wheel-ground contact points              \\ \hline
32      & Yes & center of the brand logo in the front       \\ \hline
\end{tabular}
\vspace{-0.1in}
\end{table}

\subsection{Vehicle Model Fitting}

We collected 200 vehicle 3D models from the Internet and annotated all 33 keypoints in 3D for each model. These 3D models include vehicles of various types, and an example is shown in Fig. \ref{fig:keypoints} (a). We also preprocessed the 3D models to the actual scale of real-world vehicles. For each vehicle model, we concatenate the $(x, y, z)$ coordinates of all 33 annotated 3D keypoints as a long vector (denoted as the shape vector $\mathbf{s}_i$). After that, we run Principal Component Analysis (PCA)  \cite{cootes1995active} on the set of shape vectors $\{\mathbf{s}_i\}$ of all vehicles to find the mean shape (denoted as $\mathbf{s}_{m}$), the \textit{k} basis vectors (denoted as the columns of a matrix $W$) corresponding to the \textit{k} largest eigen values, and the \textit{k}-dimensional parameter vectors $\{\mathbf{b}_i\}$, such that the reconstructed shapes $\{\hat{\mathbf{s}}_i = W\mathbf{b}_i + \mathbf{s}_{m}\}$ can approximate the original shapes $\{\mathbf{s}_i\}$. Similarly, we can generate a vehicle shape $\hat{\mathbf{s}}^{*} = W\mathbf{b}^{*} + \mathbf{s}_{m}$ from an arbitrary parameter vector $\mathbf{b}$. Shapes of vehicles of various types can be generated in this way, as demonstrated in Fig. \ref{fig:keypoints} (c). The mean shape vector $\mathbf{s}_{m}$ and the matrix $W$ are collectively called the \textbf{vehicle shape prior}.

Given a vehicle on an image, we try to find a parameter vector $\mathbf{b}$ and the vehicle pose $(R, \mathbf{t})$, such that the generated vehicle shape best fits the detected keypoints $\mathbf{p}$ under the camera projection $\Pi()$ obtained from recalibration, \textit{i.e.},
$$\underset{\mathbf{b}, R, \mathbf{t}}{\arg\min} \sum_{j}^{N}\alpha^{(j)}|| \mathbf{p}^{(j)} - \Pi(R(W^{(j)}\mathbf{b} + \mathbf{s}^{(j)}_{m}) + \mathbf{t}) || + \lambda || \mathbf{b} - \mathbf{b}_t||.$$
Here, $N$ is the total number of detected keypoints (19 in our case); $\alpha^{(j)}$ is the visibility of the \textit{jth} keypoint reported by the detector, i.e., either 1 (visible) or 0 (invisible); $\mathbf{p}^{(j)}$ is the pixel coordinates of the \textit{jth} keypoint; $W^{(j)}$ and $\mathbf{s}^{(j)}_{m}$ are the vehicle shape prior components for the \textit{jth} keypoint.

\begin{figure}[]
\vspace{0.07in}
    % \centering
    \begin{center}
        \includegraphics[width=3.35in]{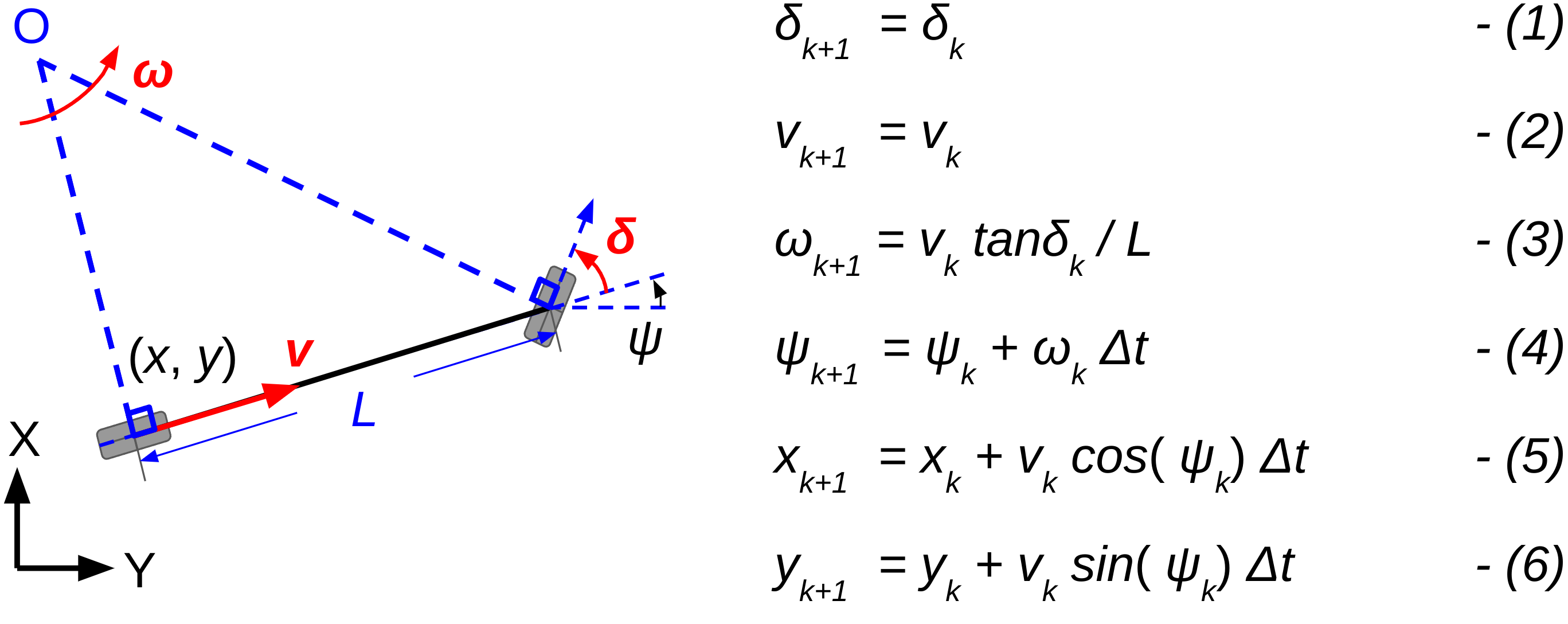}
    \caption{The simplified vehicle kinematic bicycle model (left) and state prediction rules (right).}
    \label{fig:estimator}
    \end{center}
\vspace{-0.25in}
\end{figure}

Assuming the vehicle is always on the flat ground (\textit{i.e.}, the XOY plane), there are essentially three unknown variables in the vehicle pose, \textit{i.e.}, the vehicle position $(x, y)$ in $\mathbf{t}$ and the heading angle $\psi$ in $R$ ($R$ is the rotation matrix along the z-axis by the angle $\psi$). With this parameterization, the model fitting problem is simplified to an unconstrained nonlinear least square problem, which can be solved efficiently using the Levenberg-Marquardt method \cite{more1978levenberg}. The initial position of the vehicle is approximated by the center of the bounding box, and the initial heading of the vehicle is obtained using a set of vectors through random sample consensus (RANSAC) \cite{fischler1981random}. These vectors are derived from a set of keypoint pairs pointing in the vehicle's forward direction, such as \{(0, 2), (1, 3), (4, 6), (8, 10), ...\}. In fact, since $\Pi()$ is close to a weak perspective projection for aerial videos, if the initial estimation of the vehicle heading is reasonably accurate (which is usually the case), this problem is very close to a linear least square problem. Hence, it generally converges very fast (sub-millisecond in our implementation). 

The last term $\lambda || \mathbf{b} - \mathbf{b}_t||$ is a regularizer, where $\mathbf{b}_t$ is the categorical ``template'' parameter vector. For examples, if the vehicle is detected as a sedan, $\mathbf{b}_t$ is the average of $\{\mathbf{b}_i\}$ from all sedans among the 200 vehicle 3D models that are used to construct the vehicle shape prior. Meanwhile, $\mathbf{b}_t$ is also used as the initial value of $\mathbf{b}$ in the optimization procedure. After the model fitting, we find the k-nearest-neighbor of $\mathbf{b}$ in $\{\mathbf{b}_i\}$ and use them to determine the type of the vehicle.

\begin{table*}[ht]
\vspace{0.04in}
\caption{Tracking evaluation results.}
\centering
\begin{tabular}{|c|c|c|c|c|c|c|c|c|c|c|c|c|c|c|c|}
\hline
Videos  & \begin{tabular}[c]{@{}c@{}}MOTA\\ (mask)\end{tabular} & \begin{tabular}[c]{@{}c@{}}MME\\ (mask)\end{tabular} & \begin{tabular}[c]{@{}c@{}}FP\\ (mask)\end{tabular} & \begin{tabular}[c]{@{}c@{}}FN\\ (mask)\end{tabular} & \begin{tabular}[c]{@{}c@{}}MOTA\\ (kp)\end{tabular} & \begin{tabular}[c]{@{}c@{}}MME\\ (kp)\end{tabular} & \begin{tabular}[c]{@{}c@{}}FP\\ (kp)\end{tabular} & \begin{tabular}[c]{@{}c@{}}FN\\ (kp)\end{tabular} & \#Objects & \#Images & \#Veh & \multicolumn{1}{c|}{IDE} & MT  & ML & VFP \\ \hline
track 1 & 98.1\%                                                & 512                                                  & 16                                                  & 639                                                 & 88.5\%                                              & 430                                                & 1                                                 & 11808                                             & 107140    & 29300    & 195        & 1                        & 193 & 1  & 0   \\ \hline
track 2 & 99.2\%                                                & 0                                                    & 73                                                  & 1399                                                & 90.1\%                                              & 0                                                  & 3                                                 & 17887                                             & 180438    & 42390    & 650        & 0                        & 648 & 2  & 0   \\ \hline
track 3 & 97.4\%                                                & 35                                                   & 943                                                 & 1503                                                & 89.6\%                                              & 10                                                 & 405                                               & 9681                                              & 96975     & 42796    & 498        & 2                        & 495 & 1  & 6   \\ \hline
\end{tabular}
\label{tb:tracking}
\vspace{-0.12in}
\end{table*}

\subsection{Vehicle State Estimation}

The model fitting step provides us the position and orientation of each detected vehicle on every image of the aerial video. We further run an Extended Kalman Filter (EKF) with a simplified kinematic bicycle model, as illustrated in Fig. \ref{fig:estimator}. The vehicle state prediction rules are listed in equations (1) to (6) in the figure. We assume that the vehicle maintains its steering angle and speed, i.e., equation (1) and (2). Among all the states, the position (\textit{x}, \textit{y}) and heading $\psi$ are considered directly observable, while the other three states are hidden (highlighted in red in Fig. \ref{fig:estimator}). The parameter vector $\mathbf{b}$ and the vehicle dimension are also estimated iteratively using the model fitting results in a similar way, assuming they do not change among images. The EKF approximation works well since the model fitting uncertainty is generally small and the vehicle motion between two adjacent frames is also small. Finally, the estimated states of all vehicles on all video images are exported as the vehicle trajectory data in Fig. \ref{fig:arch}.

\subsection{Implementation Details}

We built a prototype system that implements the proposed framework with a few small improvements. First, for some scenes, a piecewise flat ground model was used to better capture the uneven ground surface. The added cost is that more point correspondences are required to be annotated at carefully chosen places. Second, we augmented the camera recalibration step to a sparse monocular Simultaneous Localization and Mapping (SLAM) pipeline with key frame selection to improve the robustness. Third, we implemented a backup vehicle tracking and localization pipeline using the instance segmentation masks of vehicles, which is similar to \cite{lu2021carom}. When the keypoint detector misses a vehicle but the mask detector detects it, this backup pipeline works. Two additional estimators were implemented for the backup pipeline. When the vehicle heading can be obtained, an EKF with a point-mass and no-side-slip kinematic model is used. If the vehicle heading cannot be obtained, the Kalman Filter estimator in \cite{lu2021carom} is used.

\section{Empirical Evaluation}

We conducted several experiments to evaluate the proposed framework and our prototype implementation. First, we evaluate the vehicle detection and tracking performance with three video tracks taken from three different scenes. The results are shown in TABLE \ref{tb:tracking}, most of the metrics are from \cite{bernardin2006multiple}. Here, ``\#Veh'' is the number of vehicles in the video track. ``IDE'' is the number of vehicles with tracking ID errors. ``MT'' is the number of vehicles that are tracked for over 80\% of the time (i.e., ``mostly tracked''). ``ML'' is the number of vehicles that are tracked for less than 20\% of the time (i.e., ``mostly lost''). ``VFP'' is the number of non-vehicle objects that are wrongly tracked as vehicles (i.e., ``vehicle false positive''). A vehicle is considered ``tracked'' if it is either tracked by the proposed pipeline (using keypoints) or the backup pipeline (using masks). We only track vehicles on the traversable ground area labeled on the map, and we only assign a tracking ID to a vehicle if it can be detected and associated for at least five consecutive video images. We intentionally set a more strict score threshold for the keypoint detector so that there are less false positive and more false negatives (as shown in the ``FN (kp)'' column in TABLE \ref{tb:tracking}). In most cases, these false negatives can be handled by the backup pipeline with slight loss of vehicle localization accuracy. Overall, our prototype can track most of the vehicles correctly. Qualitative results and visualizations are available online in our GitHub repository \cite{CAROMGitHub}.

\begin{figure*}[ht]
\vspace{0.07in}
    % \centering
    \begin{center}
        \includegraphics[width=6.8in]{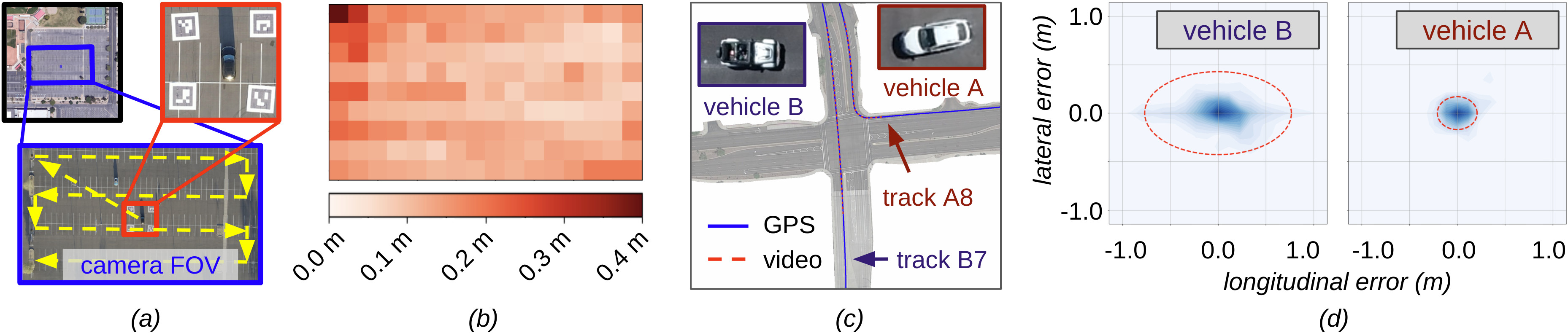}
    \caption{Empirical evaluation: (a) experiment setting for model fitting evaluation, (b) vehicle position error in the camera FOV, (c) experiment setting for vehicle tracking and localization, and (d) vehicle location error in the ego vehicle frame.}
    \label{fig:evaluation}
    \end{center}
\vspace{-0.1in}
\end{figure*}

Next, we quantitatively evaluate the model fitting performance. As shown in Fig. \ref{fig:evaluation} (a), we parked a test vehicle in an empty lot and flew the drone at 85 meters. We moved the drone in a way such that the test vehicle can be seen in different places in the camera field of view (FOV), which is illustrated as the dashed yellow trajectory in the figure. Four large ArUco markers \cite{garrido2014automatic} were placed on the ground to facilitate camera recalibration. Three different test vehicles with known dimensions are used (a sedan, a hatchback, and an SUV), and six video tracks are collected (20 minutes in total). For each video track, we marked the four contact points of the wheels and the ground to derive the ground truth vehicle pose. The evaluation results are shown in TABLE \ref{tb:evaluation_static}, averaging over all images from all video tracks. In this table, the first two data columns represent the position error in the vehicle's longitudinal direction (x) and lateral direction (y). The third data column means the error of heading angles ($\psi$) in degrees. The last three data columns are vehicle dimension errors (i.e., length, width, height) in meters. Additionally, in Fig. \ref{fig:evaluation} (b), we show the average localization error across the camera FOV. These results show that the vehicle pose and shape can be captured precisely.

\begin{table}[]
\caption{Model fitting evaluation results.}
\centering
\begin{tabular}{|l|l|l|l|l|l|l|}
\hline
Metric  & x (m) & y (m) & $\psi$ (\degree) & L (m) & W (m) & H (m) \\ \hline
Avg Error & 0.092  & 0.084  & 0.891       & 0.075       & 0.044      & 0.099       \\ \hline
Std Dev & 0.113  & 0.090  & 1.055       & 0.108       & 0.047      & 0.116       \\ \hline
\end{tabular}
\label{tb:evaluation_static}
\vspace{-0.2in}
\end{table}

\begin{figure*}[ht]
%\vspace{0.07in}
    % \centering
    \begin{center}
        \includegraphics[width=6.7in]{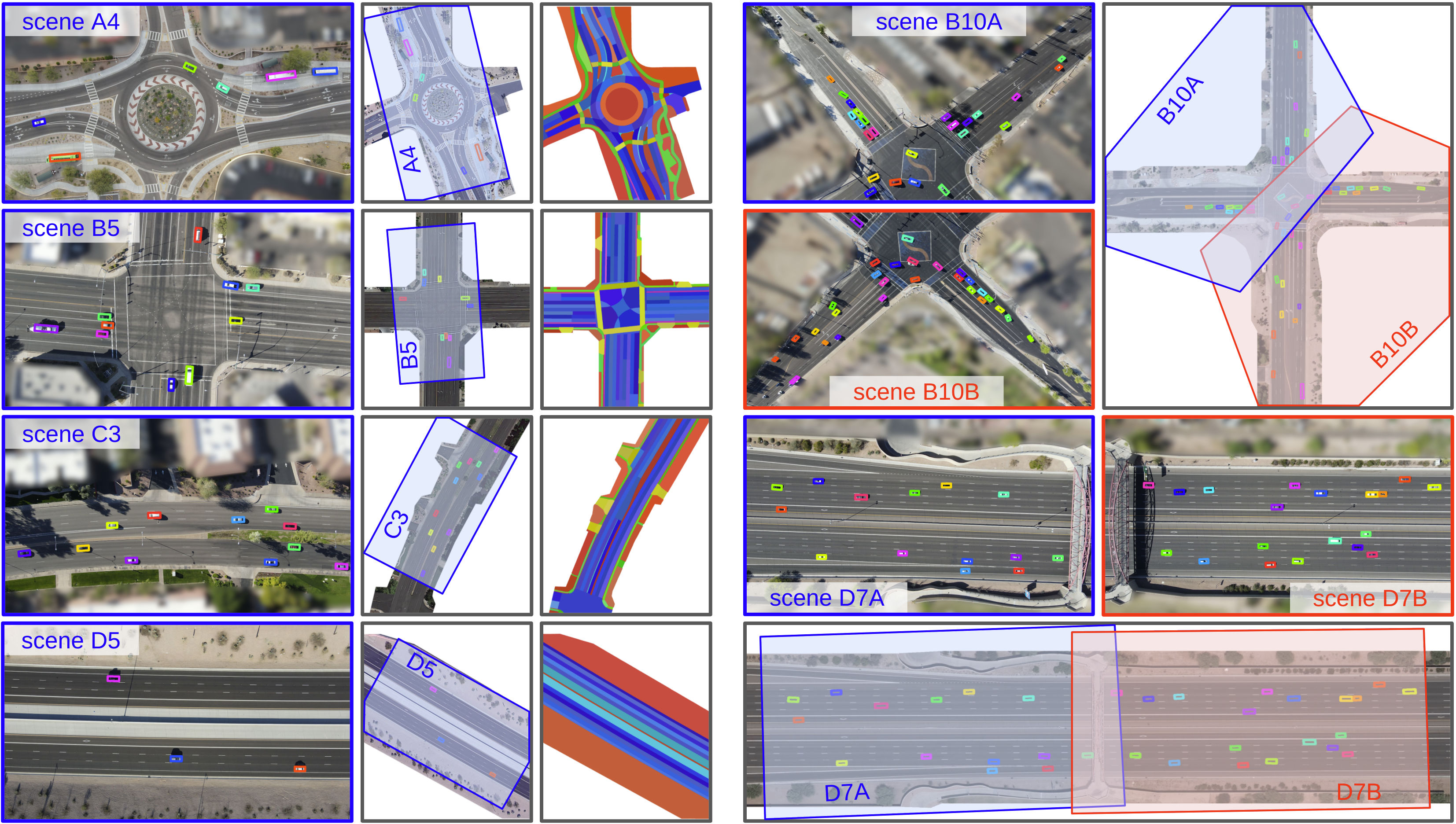}
    \caption{Examples from the CAROM Air dataset.}
    \label{fig:dataset}
    \end{center}
\vspace{-0.2in}
\end{figure*}

After that, we quantitatively evaluate the performance of vehicle localization and speed estimation. As shown in Fig. \ref{fig:evaluation} (c), we drove two test vehicles equipped with differential GPS devices in an intersection and compared the GPS data with our results obtain from a drone at 120 meters. The differential GPS has a localization accuracy of about 2 cm, and its measurements are used as references. We drove each test vehicle across the intersection 24 times in various directions. Two example trajectories are shown in Fig. \ref{fig:evaluation} (c). With our prototype system, the keypoints of vehicle A can be reliably detected all the time, and the proposed pipeline is used. In contrast, the keypoints of vehicle B can only be detected occasionally, and the backup pipeline is used most of the time. The average location difference between our method and the reference is 0.10 m and 0.26 m for vehicles A and B, respectively. The average speed estimation difference is 0.22 m/s and 0.36 m/s for vehicles A and B, respectively. Additionally, the distribution of location differences in the vehicle's reference frame is shown in Fig. \ref{fig:evaluation} (d). In this figure, the red ovals show the approximated two-sigma range, i.e., about 95\% of the measurement differences are inside the ovals. These results indicate that our framework can localize vehicles accurately. We believe that the primary sources of errors are as follows: (a) camera lens distortion, (b) inaccurate drone camera pose estimation in recalibration, (c) ground flatness, and (d) keypoint detection errors. In some cases, under strong sunlight, the detector can also make mistakes with featureless black vehicles, vehicle shadows, and the specular reflection on the vehicle surface. Besides, sometimes vehicles with similar shapes are misclassified into the wrong types, \textit{e.g.}, sedan to coupe, SUV to minivan, etc.

\begin{figure*}[ht]
\vspace{0.07in}
    % \centering
    \begin{center}
        \includegraphics[width=6.9in]{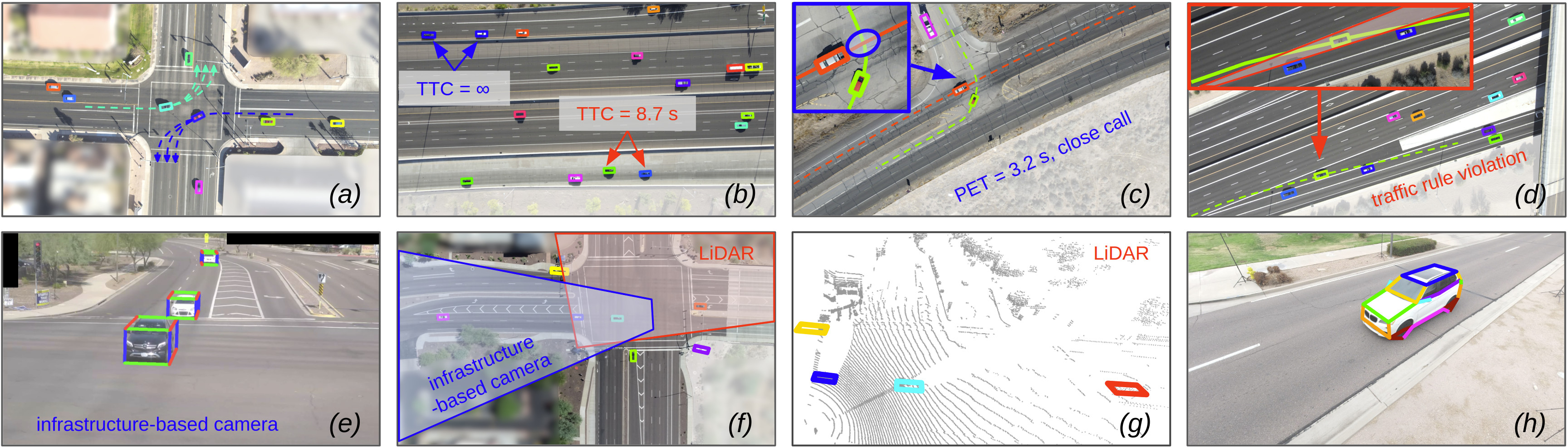}
    \caption{Example applications of the CAROM Air framework.}
    \label{fig:app}
    \end{center}
\vspace{-0.26in}
\end{figure*}

\section{The CAROM Air Dataset}

We constructed a vehicle trajectory dataset from about 100 hours of drone videos in 50 different traffic scenes covering a variety of traffic patterns, including roundabouts, intersections, local road segments, and highway segments. For a few scenes, we flew two drones simultaneously to cover larger areas, and we manually synchronize the videos using a flashlight visible from both drones. Besides the vehicle motion data, we also segmented the map at the lane level and annotated the type of these segmented areas, e.g., vehicle driving lanes, curb areas, sidewalks, crosswalks, buffer areas, etc. Examples are shown in Fig. \ref{fig:dataset}. More details about the dataset content and the data format are available online \cite{CAROMGitHub}.

\section{Applications}

In this section, we demonstrate five different applications enabled by our framework and dataset.

\textbf{(1) Fine-grained traffic counting}. Traffic counting and statistical analysis are crucial for traffic management. Our framework can automate the counting and analysis at the lane level by utilizing a semantically segmented map. Each vehicle's trajectory can be converted into a list of map segments traversed by the vehicle, and a program can count vehicles that follow a specified pattern. For example, in Fig. \ref{fig:app} (a), on the southbound (blue trajectories), we observed that the percentages of left-turning vehicles that leave the intersection in the leftmost lane, middle lane, and the rightmost lane are 45\%, 45\%, 10\%. On the northbound (cyan trajectories), the percentages are 23\%, 55\%, and 22\%. Similarly, in Fig. \ref{fig:app} (b), we can obtain the speed of vehicles on each lane using the segmented map, which shows that 54\% of the vehicles on the leftmost lane of the highway segment in both directions exceed the speed limit.

\textbf{(2) Driving safety analysis}. Various assessment metrics have been proposed to objectively evaluate driving safety \cite{wishart2020driving}\cite{jammula2022evaluation}\cite{das2023comparison}. For example, in Fig. \ref{fig:app} (b), utilizing our vehicle trajectory data and the segmented map of a traffic scene, we can compute the Time-To-Collision (TTC) metric \cite{van1993time} for any pair of adjacent vehicles in the same lane. Similarly, in Fig. \ref{fig:app} (c), given a pair of intersecting vehicle trajectories and the area of encroachment (shown as the blue-shaded circle), we can compute the Post Encroachment Time (PET) metric \cite{qi2020modified}. A low TTC or PET generally indicates unsafe driving behavior. Moreover, we can also ``re-simulate'' the motion of vehicles using our data, and then probe the safety envelope by changing the physical properties of the vehicle \cite{elli2021evaluation}.

\textbf{(3) Traffic incident detection}. Researchers spend hundreds of hours studying traffic data, which is laborious and costly. With our framework, an automated program can search through the dataset and detect incidents of interest. For example, in Fig. \ref{fig:app} (d), a vehicle drives through an area separating the main lanes on the highway and the ramp (shown as the red-shaded area). This is a traffic rule violation. We can check whether each vehicle trajectory passes through that area on the segmented map in our dataset to detect incidents of this type. Similarly, in Fig. \ref{fig:app} (c), we can detect a close call incident if the PET is lower than a threshold or an aggressive driving incident if the acceleration of a vehicle is higher than a threshold.

\textbf{(4) Reference measurement and labeling}. In order to deploy cameras and LiDARs on road infrastructure to monitor traffic, effective neural network models are needed to detect and track vehicles. However, it is expensive to construct labeled datasets to train these models, especially if it is required to label vehicle 3D bounding boxes manually. With accurate cross-sensor calibration, vehicle trajectory data generated from our framework can be used as labels for the data obtained from other sensors or as reference measurements to evaluate the performance of other traffic monitoring systems \cite{lu2021carom}\cite{srinivasan2022infrastructure}. For example, in Fig. \ref{fig:app} (f), we show the vehicle localization results on the aerial video. These results are projected onto an image obtained from an infrastructure-based camera in Fig. \ref{fig:app} (e). They are also shown in the 3D space together with the point cloud obtained from an infrastructure-based LiDAR in Fig. \ref{fig:app} (g). 

\textbf{(5) Generalization to roadside perspectives}. We can also apply our keypoint-based vehicle localization method to videos from non-aerial perspectives. An example is shown in Fig. \ref{fig:app} (h). However, the keypoint detector must be trained with data from the same perspective. If some keypoints are not observable, \textit{e.g.}, when a vehicle moves toward the camera or when it is partially occluded by another vehicle, more robust regularization is required for the model fitting step.

\section{Conclusions}

This paper presents CAROM Air, a keypoint-based vehicle localization and traffic scene reconstruction framework using aerial videos recorded by drones. Our framework achieves decimeter-level localization accuracy and enables many practical downstream traffic analysis applications. Still, it has certain limitations, such as short flight time, restricted fly zones in cities, potential risks of drone crashes, etc. The drone camera also has a limited dynamic range, and the detector can produce errors on certain vehicles that appear infrequently in our training data (e.g., motorcycles, trucks, and trailers). With further development, we hope it can serve as a flexible method for road traffic analysis and eventually help improve road safety and transportation efficiency.

\bibliographystyle{IEEEtran}
\bibliography{references}

\end{document}